\documentclass[conference, a4paper]{IEEEtran}
\usepackage{cite}
\usepackage{amsmath,amssymb,amsfonts}
\usepackage{algorithmic}
\usepackage{graphicx}
\usepackage{textcomp}
\usepackage{url}
\usepackage{multirow}
\usepackage[utf8]{inputenc} % Kullanılan encodinge göre utf8 yerine latin5 de yazılabilir.
\usepackage{paralist}
\usepackage{graphics} %% add this and next lines if pictures should be in esp format
\usepackage{epsfig} %For pictures: screened artwork should be set up with an 85 or 100 line screen
\usepackage{threeparttable}
\usepackage{epstopdf}%This is to transfer .eps figure to .pdf figure; please compile your paper using PDFLeTex or PDFTeXify.
\usepackage[colorlinks=true,bookmarks=false]{hyperref}
\usepackage{subfig} 
\usepackage{lipsum} 
\usepackage{hyphenat}
\usepackage{rotating}
\usepackage{makecell}

\AtBeginDocument{%
  \renewcommand\tablename{TABLO}
}

\AtBeginDocument{%
  \renewcommand\abstractname{Abstract}
}
\begin{document}

\title{Kısa Konuşma Cümlelerinin Dönüştürücü Yöntemleriyle  Otomatik Etiketlenmesi \\
Auto-tagging of Short Conversational Sentences using Transformer Methods}

\author{
\IEEEauthorblockN{ D. Emre Taşar\IEEEauthorrefmark{1}, Şükrü Ozan\IEEEauthorrefmark{2}, Umut Özdil\IEEEauthorrefmark{2}, M. Fatih Akca\IEEEauthorrefmark{3},Oğuzhan Ölmez\IEEEauthorrefmark{4}, \\Semih Gülüm\IEEEauthorrefmark{5}, Seçilay Kutal\IEEEauthorrefmark{5}, Ceren Belhan\IEEEauthorrefmark{6}}
 
\IEEEauthorblockA{\IEEEauthorrefmark{1} emretasa@garantibbva.com.tr}
\IEEEauthorblockA{\IEEEauthorrefmark{2} sukruozan@adresgezgini.com,umutozdil@adresgezgini.com}
\IEEEauthorblockA{\IEEEauthorrefmark{3} mehmet.akca1@ogr.sakarya.edu.tr}  
\IEEEauthorblockA{\IEEEauthorrefmark{4} 172803041@ogr.cbu.edu.tr}  
\IEEEauthorblockA{\IEEEauthorrefmark{5} semihgulum@marun.edu.tr, secilaykutal@marun.edu.tr}  
\IEEEauthorblockA{\IEEEauthorrefmark{6} ceren.belhan@std.ieu.edu.tr}  
}

\maketitle
\IEEEpubidadjcol

\hyphenation{soh-bet i-ler-le-yen  he-def-le-mek-te-yiz prob-le-mi-ni mo-del-le-ri op-tical net-works semi-conduc-tor e-ti-ket-le-di}

\begin{ozet}
Kısa konuşma cümlelerinin anlamsal özelliklerine göre yüksek doğruluk ile kategorilere ayrıştırılması problemi, doğal dil işleme alanında üzerinde çalışılan bir konudur. Bu çalışmada 46 farklı kategoride sınıflandırılan örnekler ile oluşturulan bir veri seti kullanılmıştır. Örnekler, bir firmanın müşteri temsilcileriyle, firmanın internet sitesi ziyaretçileri arasında gerçekleşen yazılı sohbet (chat) görüşmelerinden alınmış cümlelerden oluşmaktadır. Ana amaç, sorulan sorulara anlamlı yanıtlar üretebilen bir sohbet uygulamasında kullanmaya yönelik olarak internet sitesi ziyaretçilerden gelen soruları ve talepleri önceden belirlenen 46 kategori için en doğru şekilde otomatik olarak etiketlemektir. Bunun için Türkçe dilinde ön eğitime tabi tutulmuş birbirinden farklı BERT modelleri ve  bir adet GPT-2 modeli tercih edilmiştir. İlgili modellerin sınıflandırma başarımları detaylı bir şekilde incelenerek raporlanmıştır.
\end{ozet}

\begin{IEEEanahtar}
otomatik etiketleme, doğal dil işleme, BERT, GPT-2, çoklu sınıf.
\end{IEEEanahtar}

\begin{abstract}
The problem of categorizing short speech sentences according to their semantic features with high accuracy is a subject studied in natural language processing. In this study, a data set created with samples classified in 46 different categories was used. Examples consist of sentences taken from chat conversations between a company's customer representatives and the company's website visitors. The primary purpose is to automatically tag questions and requests from visitors in the most accurate way for 46 predetermined categories for use in a chat application to generate meaningful answers to the questions asked by the website visitors. For this, different BERT models and one GPT-2 model, pre-trained in Turkish, were preferred. The classification performances of the relevant models were analyzed in detail and reported accordingly.
% \boldmath
\end{abstract}
\IEEEpeerreviewmaketitle
\IEEEpubidadjcol

\section{G{\footnotesize İ}r{\footnotesize İ}ş}\label{sec:giris}

Bu çalışmada doğal dil işleme modelleri yardımı ile sorulara otomatik olarak yanıtlar üretmesi hedeflenen bir chatbot için, gelen soruları önceden belirlenmiş 46 temel kategoriye mümkün olduğunca doğru bir şekilde ayırabilmek hedeflenmektedir. Günümüzde yoğun olarak çalışılmakta olan bu konu ile ilgili olarak literatürde birçok güncel çalışmaya rastlamak mümkündür.

Metin sınıflandırma problemi için Jain vd. tarafından hibrit kümeleme ve sınıflandırma modeli (HCC)\cite{jain2021building} önerilmiştir. Bu modelde, metinlerin kümelenmesi için k-ortalama metodu işleme alınırken, çok etiketli bir sınıflandırma yapılabilmesi için ise derin öğrenme algoritmaları kullanılmıştır. Çalışma sonunda sınıflandırma problemi için Evrişimli Sinir Ağları (CNN) başarılı bir sonuç verirken, çok etiketli sınıflandırma için  Uzun-Kısa Süreli Bellek (LSTM) ya da Derin CNN'in daha etkili olabileceği belirtilmiştir.

BERT modeli doğal dil işlemede metnin sınıflandırması için kullanılan açık kaynaklı bir modeldir. Devlin vd. tarafından geliştirilen transformatörlerden çift yönlü kodlayıcı gösterimlere sahip olan BERT modeli \cite{Devlin:2018} ile birlikte, etiketlenmemiş metinlerin derin çift yönlü temsillerinin önceden eğitimi sırasında tüm katmanlarda sağ ve sol bağlam bilgilerinin dahil edilmesi sağlanmaktadır. Model, sonrasında ince ayar yapılarak göreve özgü şekilde eğitilebilmektedir. Bu görevler soru cevaplama, duygu analizi, metin sınıflandırma ve adlandırılmış varlık tanıma gibi farklı çeşitlerde olabilir\cite{ozccift2021advancing}.

İnce ayar yapılabilmesi için önceden eğitilmiş çeşitli BERT modelleri mevcuttur. Bu modeller dil açısından, dile özgü ve çok dilli olarak ikiye ayrılmaktadır. Türkçe dili özelinde literatürde yeteri kadar iyi eğitilmiş çok fazla dil modeli olmadığı için çalışmada  Türkçe diline özel olarak eğitilmiş olanların yanı sıra çok dilli modeller de kullanılmıştır. Ronnqvist vd. çalışmalarında \cite{ronnqvist2019multilingual} kullandıkları çok dilli model ile 6 dil üzerinde çeşitli görevlerde test gerçekleştirmiş ve bu modelin başarısının diller arasındaki değişimini gözlemlemişlerdir. Çalışmada Türkçe ile aynı dil ailesine üye olan Fince dilinde model \%93.2 doğruluk değerine ulaşmıştır.

Özçift vd. çalışmasında \cite{ozccift2021advancing} morfolojik olarak zor bir dil olan Türkçe ile 6 farklı görevde geleneksel makine öğrenmesi yöntemlerinin yanında BERT modelinin performansını karşılaştırmıştır. Spam etiketleme görevi dışında, BERT modeli diğer modellere nazaran daha yüksek doğruluk oranı vermiştir.

Tanaka vd. çalışmasında\cite{tanaka2019document} BERT modelini sınıflandırıcı olarak kullanmıştır. Dokümanların sınıflandırılması için ince ayar yapılan BERT modelinde, girdi olarak doküman verilmiş ve BERT'in metin sınıflandırmadaki başarısı ortaya konulmuştur.
 
Ozan vd. çalışmasında \cite{adresgezgini} kategorik sınıflandırma probleminde yaygın olarak kullanılan Doc2Vec, LSTM ve BERT gibi modellerin performanslarını karşılaştırmıştır. Çalışmada BERT modelinin \%93 doğruluk seviyesiyle diğer modellere göre yüksek başarım gösterdiği tespit edilmiştir.

Bahsedilen çalışmalar doğrultusunda seçilen BERT modelleri, Bölüm \ref{sec:data}'de detayları verilen veri seti ile, sınıflandırma problemine yönelik olarak  ince ayar yapılarak eğitilmiştir.

BERT modelleri dışında çalışmada yer verilen bir diğer model olan GPT-2, Radford vd. tarafından geliştirilmiş üretken yapıya sahip olan bir kod çözücü transformatördür \cite{radford2019language}.  Çalışmada bu model, kendi içinde özel olarak sınıflandırma yapan bir kütüphanenin yardımı ile metin sınıflandırma amacıyla kullanılmıştır.

Whitfield, sınıflandırma problemlerinde model performansını arttırabilmek adına GPT-2 modelini kullanarak ürettiği sentetik verilerle, veri setini büyüterek başarım metriklerini karşılaştırmayı hedeflemiştir \cite{whitfield2021using}. Çalışmadan ulaşılan sonuçlara göre GPT-2, kuruluşların ve işletmelerin, büyük ölçekli veri toplamaya ilişkin yüksek maliyetlere katlanmadan yüksek performansa sahip sınıflandırma modelleri oluşturmasına izin vermektedir.

\section{Kullanılan Ver{\footnotesize İ} Set{\footnotesize İ}} \label{sec:data}

\begin{figure}[h!]
  \centering
        \includegraphics[width=0.45\textwidth]{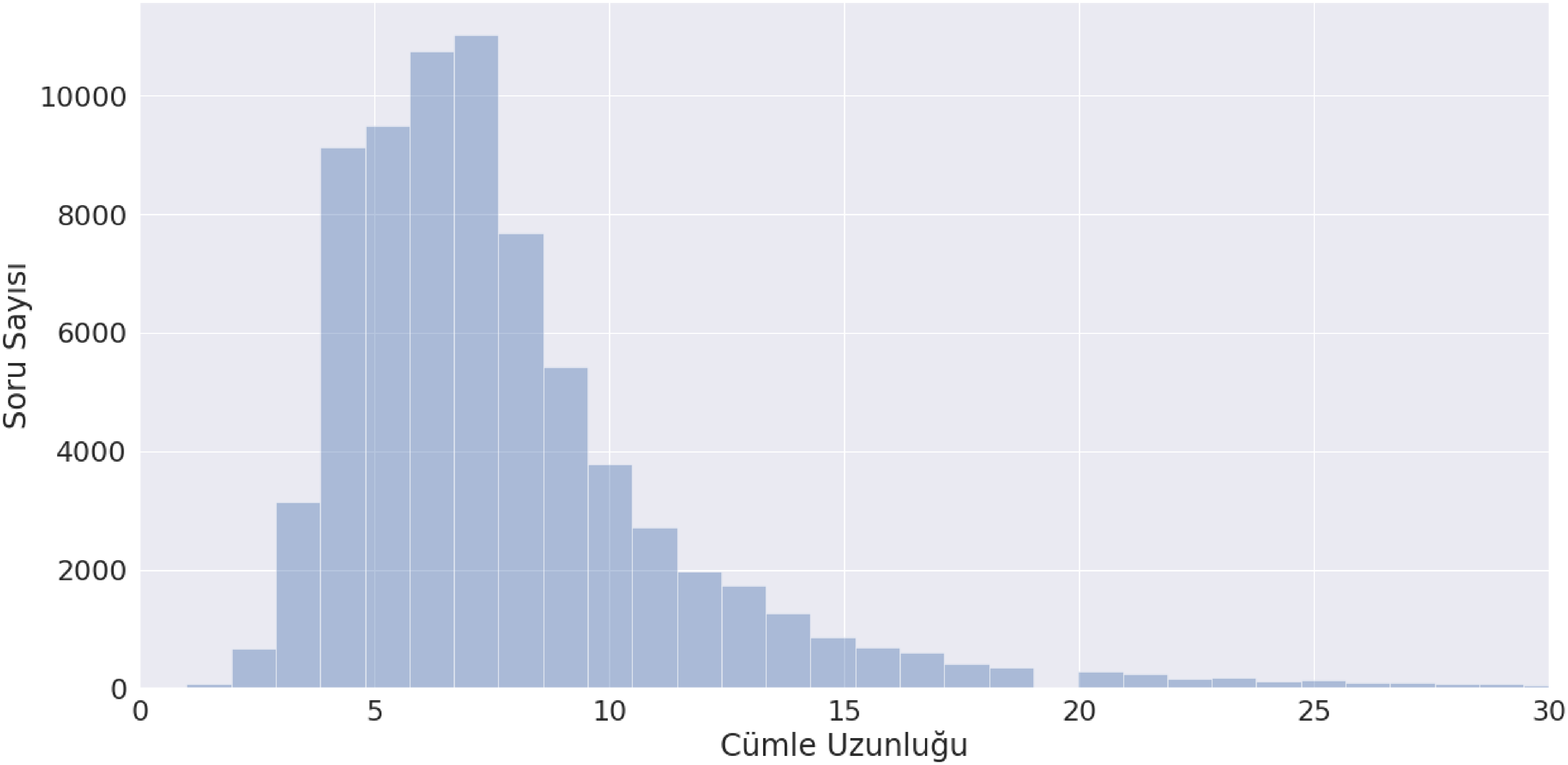}
        \caption{Veri Setindeki Cümlelerin Uzunluklarına Göre Dağılım Grafiği}
        \label{img:distplot}
        \includegraphics[width=0.45\textwidth]{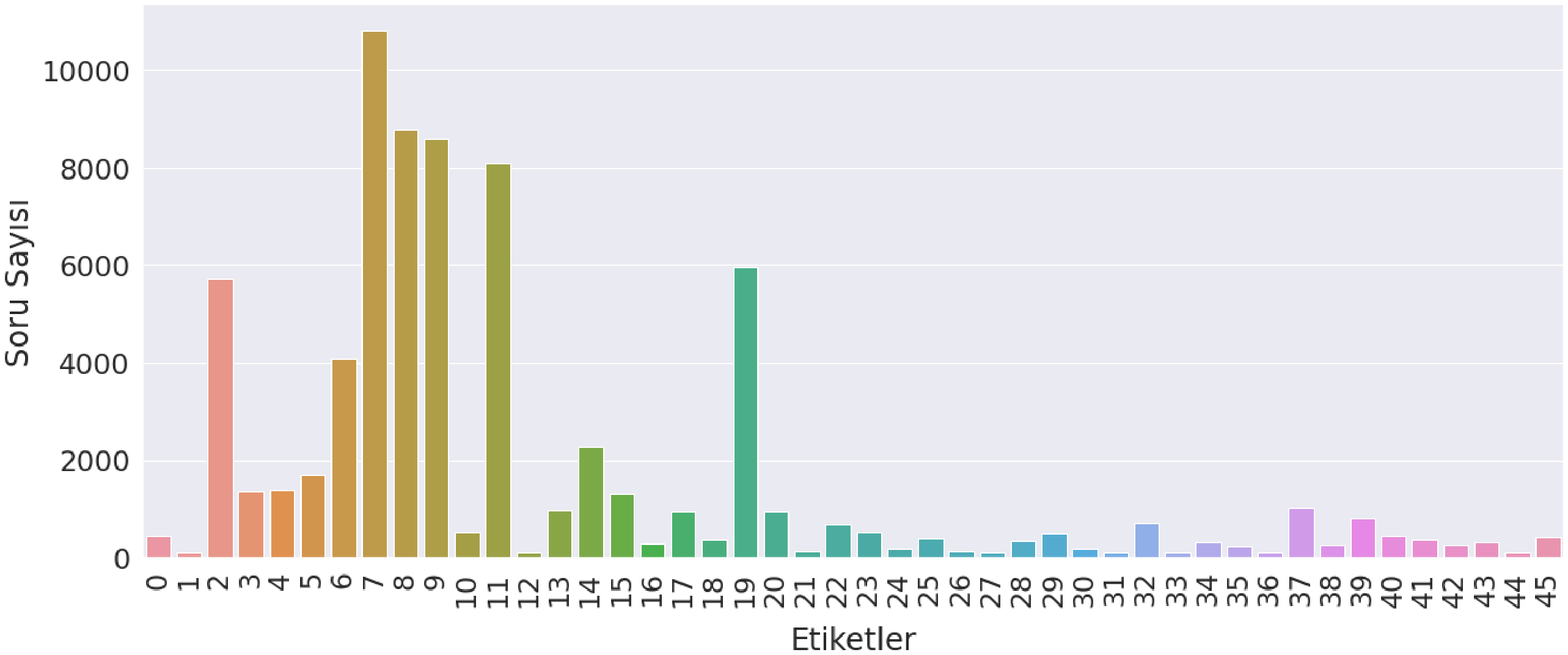}
        \caption{Veri Setinin Etiketlere Göre Dağılım Grafiği}
        \label{img:etiketveridagilimi}
\end{figure}

Veri seti için firmanın internet sitesinde kullanılan sohbet arayüzü aracılığıyla web sitesi ziyaretçileri tarafından yöneltilen sorular ele alınmıştır. Elde edilen 73551 adet ve kelime sayısı 2 ile 30 arasında değişen seçili cümle, 46 kategori için manuel olarak etiketlenmiştir. Örnek yedi kategori için seçilmiş bazı örnek cümleler Tablo \ref{tab:ornekler}’de verilmiştir. Veri setinde bulunan cümle-cümle gruplarının uzunluğunun geniş bir aralıkta değişkenlik göstermesi (Şekil \ref{img:distplot}), kategorilere göre veri dağılımının düzensiz oluşu (Şekil \ref{img:etiketveridagilimi}) ve benzer kalıplarda verilerin olması hazırlanan veri setinin kullanılmasının zorlukları olarak sayılabilir. Ön işleme aşamasında veri setindeki cümlelerde yer alan sayılar ve noktalama işaretleri atılmış, bütün harfler küçük harflere çevrilmiştir. Veri seti \%70'i eğitim, \%30'u da test için olacak şekilde iki gruba ayrılmıştır. Eğitim aşamasında kullanılan yöntemlere ait detaylar Bölüm \ref{sec:methods}’de verilecektir.

%Örnek Cümleler ve Eşleştirilmiş Etiketler tablosu
\begin{table}[h!]
\caption{Örnek Cümleler ve Eşleştirilmiş Etiketler}
\centering
\begin{tabular}{|l|l|} 
\hline

\multirow{3}{*}{\begin{tabular}[c]{@{}l@{}}Facebook\\reklamları\end{tabular}} & Facebook reklamı için görüşmek istiyorum \\ 
\cline{2-2} & \begin{tabular}[c]{@{}l@{}}Facebook reklam paketleri, ücretleri ve\\kampanyalarınız ile ilgili bilgi verir misiniz?\end{tabular}   \\ 
\cline{2-2} & sitemi Facebook reklam yapmak istiyoreum  \\ 
\hline
\multirow{3}{*}{\begin{tabular}[c]{@{}l@{}}Instagram\\reklamları\end{tabular}} & \begin{tabular}[c]{@{}l@{}}Instagram şirket reklamı vermek istiyordum\\bu konu ile yardımcı olabilir misiniz\end{tabular}   \\ 
\cline{2-2} & Instagram nasıl reklam verebilirim  \\ 
\cline{2-2} & Instagram reklamlarıyla ilgili bilgi almak istemiştim  \\ 
\hline

\multirow{3}{*}{\begin{tabular}[c]{@{}l@{}}Mobil\\uygulama\end{tabular}} & \begin{tabular}[c]{@{}l@{}}mobil uygulamalarının web entegrasyonunu yapabiliyor\\musunuz\end{tabular}\\ 
\cline{2-2} & \begin{tabular}[c]{@{}l@{}}Merhaba, mobil uygulama için fiyat teklifi almak istiyorum\\detayları hangi mail adresi ile paylaşabilirim?\end{tabular}   \\ 
\cline{2-2} & mobil uygulama yaptırmak istiyorum ne yapmalıyım  \\ 
\hline

\multirow{3}{*}{\begin{tabular}[c]{@{}l@{}}Ofisiniz\\nerede\end{tabular}} & Adres bilgisi alabilir myim  \\ 
\cline{2-2} & hangi ilden hizmet veriyorsunuz  \\ 
\cline{2-2} & \begin{tabular}[c]{@{}l@{}}Ankarada yüzyüze görüşüp  ilgi alabileceğimiz bir ekibiniz \\mevcut mudur\end{tabular}  \\ 
\hline

\multirow{3}{*}{\begin{tabular}[c]{@{}l@{}}İş\\başvurusu\end{tabular}} & firmanızda çalışma imkanımız var mı  \\ 
\cline{2-2} & \begin{tabular}[c]{@{}l@{}}Bu yaz için staj imkanlarınızı öğrenmek istiyorum. \\Bilgisayar Mühendisliği 3.sınıf öğrencisiyim\end{tabular} \\ 
\cline{2-2} & acaba iş başvuru yapabilmesi için mail adresiniz var mıdır  \\ 
\hline

\multirow{3}{*}{\begin{tabular}[c]{@{}l@{}}Web sitesine canlı\\destek ekleme\end{tabular}} & Canlı destek hizmeti sunuyormusunuz  \\ 
\cline{2-2} & \begin{tabular}[c]{@{}l@{}}müşteri telefonlarına yetişemiyoruz canlı\\destek hizmetinizden yararlanmak isteriz ücreti ne kadar\end{tabular} \\ 
\cline{2-2} & Birde bu canlı destek sisteminizin ücreti nedir ?  \\ 
\hline

\multirow{3}{*}{\begin{tabular}[c]{@{}l@{}}Web sitesi\\yaptırmak\end{tabular}} & Siz web tasarim yapiyrmusunz peki  \\ 
\cline{2-2} & ben web sitesi yaptırmayı düşünüyorum \\ 
\cline{2-2} & ya peki web adresi kurabilir  miyiz  \\ 
\hline
\end{tabular}
\begin{tablenotes}
   \item Not: Toplam kategori sayısı 46 olmasına karşın tabloda 7 örnek kategori paylaşılmıştır.
\end{tablenotes}
\label{tab:ornekler}
\end{table}

%Veri Setindeki Cümlelerin Uzunluklarına Göre Dağılım Grafiği

\section{Yöntemler} \label{sec:methods}

\subsection{BERT Yöntemi} \label{subsec:bert}

BERT modeli ön eğitim ve ince ayar olarak 2 aşamadan meydana gelmektedir. Ön eğitim, birbirinden farklı ve etiketlenmemiş veriler üzerinden bir konuşma diline özel yada birden fazla konuşma dili için eğitilir. Ön eğitimden geçmiş bir BERT modeline etiketlenmiş veriler ile ince ayar yapılır. İnce ayar sayesinde BERT'in dikkat mekanizması neye dikkat etmesi gerektiğini anlayabilir. Çalışmada kullanılan BERT-base modeli, 12 transformatör bloğu, 12 öz-dikkat başlığı ve 768 gizli katmana sahip bir kodlayıcı içerir ve çift yönlü öz-dikkat kullanır \cite{Sun2019HowTF}. Kullanılan distil-BERT modeli ise BERT-base modeline göre daha küçük bir yapıya sahip olduğundan \%60 daha hızlı çalışırken BERT performansını da büyük ölçüde koruyabilmektedir \cite{sanh2019distilbert}. BERT modelinin eğitimi büyük harflere duyarlılık açısından, "cased" (büyük küçük harf kullanımına karşı duyarlı) ve "uncased" (büyük küçük harf kullanımına karşı duyarsız) olarak iki farklı model tipi ile gerçekleştirilebilmektedir.

\subsection{GPT-2 Yöntemi} \label{subsec:gpt2}

Mevcut sistemler, genel veri setleri ile çalışmak yerine daha dar konularda çalışmaya uygundur. ``Üretken Eğitim Öncesi Transformatör" olarak adlandırılan ve anlamlı metin oluşturmada oldukça başarılı olan GPT-2,  mevcut sistemlerin aksine çoklu görevlerde çalışabilecek ve özel görevlere göre de konfigüre edilebilecek şekilde tasarlanmıştır.

Bir kod çözücü transformatörü olan GPT-2'nin eğitimi esnasında girdi dizisinin son belirteci, tahminde ihtiyaç duyulan tüm bilgileri içerdiğinden, girdiyi takip eden bir sonraki token hakkında tahmin yapılması amacıyla kullanılmaktadır. Girdi dizisinin son belirtecinin taşıdığı bilgi, üretim problemi yerine bir sınıflandırma probleminde tahmin yapmak amacıyla da kullanılabilir. GPT-2 ile tahmin yapılırken, BERT modelinin sınıflandırma probleminde uygulandığı gibi ilk token yerleştirmek yerine son token yerleştirme uygulanmaktadır \cite{mihaila2021}.

Bu kapsamda gerçekleştirilmeye çalışılan sınıflandırma problemi için Bölüm \ref{sec:giris}’de bahsedilen yöntemler ile birlikte BERT ve GPT-2 modellerinin başarımı test edilmiştir. Daha önceden firma bünyesinde yapılmış olan çalışma \cite{adresgezgini} genişletilerek toplamda 10 adet BERT modeli ve bir adet GPT-2 modeli kullanılmıştır. GPT-2'nin sınıflandırıcı olarak kullanıldığı çalışmada, ilgili modeller söz konusu veri seti ile ince ayar yapılarak eğitilmiştir. Model mimarilerine ait bazı temel parametrelere ait değerler Tablo \ref{tab:model}'de gösterilmiştir.

\subsection{F1 Skoru ve Başarım Kriterleri} \label{subsec:metric}

Eğitilen modellerin başarısı doğruluk değeri, F1 skoru, kayıp değerleri ve hata dizeyi (confusion matrix) baz alınarak değerlendirilmiştir. Bu başarı ölçütleri; True Positive (TP), False Positive (FP), True Negative (TN) ve False Negative (FN) değerlerine bakılarak hesaplanır. Modelin çıktısı ikili sınıflandırma olarak göz önüne alınırsa; TP, modelin tahmini bir durum için olumlu sonuç vermesi ve asıl durumda da sonucun olumlu olduğunu gösterirken TN, algoritma çıktısının olumsuz iken asıl durumun da olumsuz olduğunu temsil eder. İki tip hata vardır, bunlar FP ve FN kavramlarıdır.FP algoritmanın tahminini olumluyken asıl değerin olumsuz olması durumudur. FN ise algoritmanın olumsuz dediği bir sonuç karşısında asıl durumun olumlu olmasıdır. Çalışmada kullanılan veri setinde örnek bir etiket için  elde edilen sonuçlarla oluşturulmuş olan Tablo \ref{tab:confmatrix}, TP, FP, TN ve FN kavramlarının daha iyi anlaşılabilmesi için incelenebilir.

%Kullanılan Ön Eğitimli BERT Modelleri ve GPT-2 Modelinin Parametreleri
\begin{table*}[h!]
\centering
\caption{Kullanılan Ön Eğitimli BERT Modelleri ve GPT-2 Modelinin Parametreleri}
\begin{tabular}{|l|c|c|c|c|l|c|c|} 
\hline
\textbf{Model} & \begin{tabular}[c]{@{}c@{}}\textbf{Gizli}\\\textbf{Katman}\\\textbf{Boyutu}\end{tabular} & \begin{tabular}[c]{@{}c@{}}\textbf{Maksimum}\\\textbf{Sekans}\\\textbf{Uzunluğu}\end{tabular} & \begin{tabular}[c]{@{}c@{}}\textbf{Dikkat Ana}\\\textbf{Başlığı}\\\textbf{Sayısı}\end{tabular} & \begin{tabular}[c]{@{}c@{}}\textbf{Gizli}\\\textbf{Katman}\\\textbf{Sayısı}\end{tabular} & \textbf{Mimari} & \textbf{Dil} & \begin{tabular}[c]{@{}c@{}}\textbf{Sözlük}\\\textbf{Boyutu}\end{tabular}  \\ 
\hline
bert-base-multilingual-cased \cite{DBLP:journals/corr/abs-1810-04805} & \multirow{10}{*}{768}                                                                                                                & \multirow{10}{*}{512}
                                                                      & \multirow{10}{*}{12} 
                                                                      & \multirow{9}{*}{12} 
                                                                      & \multirow{9}{*}{BertForMaskedLM} & Çok dilli & 119547 \\ 
\cline{1-1}\cline{7-8}
bert-base-multilingual-uncased \cite{DBLP:journals/corr/abs-1810-04805} & & & & & & Çok dilli & 105879 \\ 

\cline{1-1}\cline{7-8}
dbmdz/bert-base-turkish-128k-cased \cite{dbmdzbertbaseturkish128kcased}   & & & & & & Türkçe & 128000 \\ 

\cline{1-1}\cline{7-8}
dbmdz/bert-base-turkish-128k-uncased \cite{dbmdzbertbaseturkish128kuncased} & & & & & & Türkçe & 128000 \\ 

\cline{1-1}\cline{7-8}
dbmdz/bert-base-turkish-cased \cite{dbmdzbertbaseturkishcased02} & & & & & & Türkçe & 32000 \\ 

\cline{1-1}\cline{7-8}
dbmdz/bert-base-turkish-uncased \cite{dbmdzbertbaseturkish128kuncased} & & & & & & Türkçe & 32000 \\ 

\cline{1-1}\cline{7-8}
Geotrend/bert-base-tr-cased\cite{smallermbert} & & & & & & Türkçe & 19099 \\ 

\cline{1-1}\cline{7-8}
loodos/bert-base-turkish-cased \cite{loodosbertbaseturkishcased} & & & & & & Türkçe & 32000 \\ 

\cline{1-1}\cline{7-8}
loodos/bert-base-turkish-uncased \cite{loodosbertbaseturkishununcased} & & & & & & Türkçe & 32000 \\ 

\cline{1-1}\cline{5-8}
dbmdz/distilbert-base-turkish-cased \cite{dbmdzdistilbertbaseturkishcased} & & & & 6 & DistilBertForMaskedLM & Türkçe & 32000\\ 
\hline
adresgezgini/turkish-gpt-2 \cite{AdresGezgini2} & 768 & 1024 & 12 & 12 & GPT2LMHeadModel & Türkçe & 50000 \\
\hline
\end{tabular}
\label{tab:model}
\end{table*}

%karmaşıklık matrisi tablosu
\begin{table}[h!]
\centering
\caption{Karmaşıklık Matrisi}
\begin{tabular}{|l|c|c|c|} 
\cline{3-4}
\multicolumn{1}{c}{} & & \multicolumn{2}{c|}{Ait olduğu sınıf} \\ 
\cline{3-4}
\multicolumn{1}{c}{\begin{sideways}\end{sideways}}                             & & İş başvurusu (1) & İş başvurusu değil (0) \\ 
\hline
\multirow{2}{*}{\begin{tabular}[c]{@{}c@{}}Tahmin\\edilen\\sınıf\end{tabular}} & \begin{tabular}[c]{@{}c@{}}İş başvurusu\\(1)\end{tabular}       & \begin{tabular}[c]{@{}c@{}}Şirketinizde çalışmak için\\nereyi doldurmam gerekir\\\textbf{TP}\end{tabular} & \begin{tabular}[c]{@{}c@{}}İşyerim için reklam\\vermek istiyorum\\\textbf{FP}\end{tabular}         \\ 
\cline{2-4} & \begin{tabular}[c]{@{}c@{}}İş başvurusu\\değil (0)\end{tabular} & \begin{tabular}[c]{@{}c@{}}İşe alımlar devam\\ediyor mu\\\textbf{FN}\\\end{tabular} & \begin{tabular}[c]{@{}c@{}}Sosyal medya reklamları\\nasıl oluşturuluyor\\\textbf{TN}\end{tabular}  \\
\hline
\end{tabular}
\label{tab:confmatrix}
\end{table}

Doğruluk değeri en temel başarım kriteri olarak ifade edilebilir. Bu metrikte modelin ürettiği doğru tahmin sonuçları üretilen tüm tahminlere oranlanmaktadır (Denklem \ref{eq:accuracy}). Ancak bu yöntem kategorik sınıflandırma problemlerinde sınıf bazlı bir sonuç veremediğinden tercih edilmemektedir. Bu durumlarda F1 skoru adı verilen ve kesinlik-doğruluk değerlerine dayanan ayrı bir metrik kullanılmaktadır.

Kesinlik (precision) değeri modelin tahmin ettiği TP sayısının modelin ürettiği tüm olumlu sonuçlara oranı ile bulunmaktadır (Denklem \ref{eq:precision}). Duyarlılık (recall) değeri ise modelin tahmin ettiği TP sayısının modelin tahmin etmesi gereken tüm olumlu sonuçlara oranı ile bulunmaktadır (Denklem \ref{eq:recall}). Bu iki kavram modelin ürettiği sonuçların farklı iki durumu üzerinden hesaplandığından her iki durumun da etkin bir şekilde yorumlandığı F1 skoru, Denklem \ref{eq:f1score}'te ifade edildiği gibi kesinlik ve duyarlılık değerlerinin harmonik ortalaması alınarak hesaplanmaktadır\cite{powers2020evaluation}.

%Formüller
\begin{equation} \label{eq:accuracy}
    Do\Breve{g}ruluk =  \frac{TP+TN}{TP+TN+FP+FN}
\end{equation}

\begin{equation} \label{eq:precision}
    Kesinlik =  \frac{TP}{TP+FP}
\end{equation}

\begin{equation} \label{eq:recall}
    Duyarlılık = \frac{TP}{TP+FN}
\end{equation}

\begin{equation} \label{eq:f1score}
    F1 skoru = 2\left (  \frac{Kesinlik\times Duyarlılık}{Kesinlik+Duyarlılık}\right )
\end{equation}

Kategorilere göre veri sayısının düzenli dağılmaması, çalışmada kullanılan modellerin genel başarımının tespiti için farklı bir metriğin kullanımını gerektirmektedir. Kategori bazında hesaplanmış F1 skorunun kategori bazlı veri dağılımına göre ağırlıklı ortamasının alınarak hesaplanmasıyla dengesiz veri dağılımında ortaya çıkan bu problem belirli bir ölçüde çözülebilmektedir.

%Makine öğrenmesi ve derin öğrenme modellerinde hata oranını minimize etmek için kayıp fonksiyonundan yararlanılır. Kayıp fonksiyonu hataları minimize etmek için geriye dönük olarak çalışan bir optimizasyon sürecidir\cite{goodfellow2016deep}. Eğitim süresince her dönem sonrasında ortaya çıkan kayıp değeri, modelin eğitim dönemi sırasındaki kayıp fonksiyonundan çıkan son değerini ifade eder. Bu değer ile modelin başarı oranı ters orantılıdır.

\section{Uygulama Sonuçları}

\begin{figure*}[h!]
	\centering
		\includegraphics[width=0.7\textwidth]{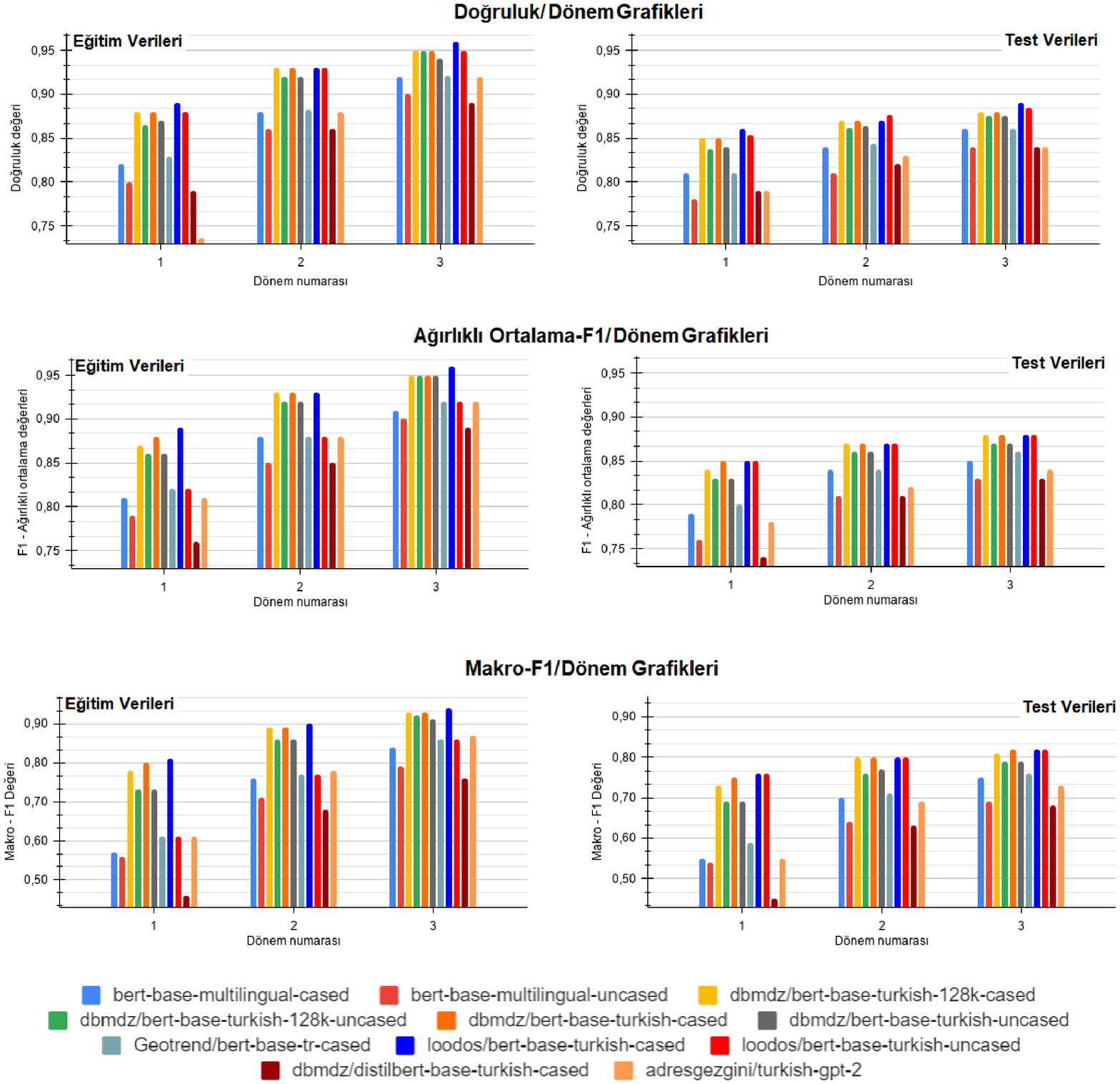}
	\caption{BERT ve GPT-2 Modelleri Eğitim Sonuçları}
	\label{img:sonucgrafik}
\end{figure*}

Çalışmada sınıflandırma problemi, Bölüm \ref{subsec:bert} ve Bölüm \ref{subsec:gpt2}'de anlatıldığı gibi farklı BERT-base, distilBERT-base ve GPT-2 modellerinin eğitimleri ile çözülmeye çalışılmıştır. Bu bölümde ise eğitilen modellerin sonuçları Bölüm \ref{subsec:metric}'de verilen kriterler göz önüne alınarak incelenecektir. Her modelin eğitimi üç dönem (epoch) gerçekleştirilerek parametreler hesaplanmıştır. Şekil \ref{img:sonucgrafik}'de modellerin doğruluk/dönem grafikleri çizdirilmiştir.

GPT-2 metin üretmede göstermiş olduğu başarısını metin sınıflandırma probleminde gösterememiş ve doğruluk değeri ancak \%84 değerine kadar çıkabilmiştir. Ancak çalışma için daha önemli bir karşılaştırma kriteri olan F1 skoru \%73 olarak gözlemlenmiştir.

BERT modeli için önceden eğitilmiş ``cased'' ve ``uncased'' modeller ile gerçekleştirilen ince ayar eğitimlerinin ardından başarım oranları incelendiğinde, iki farklı model tipinin sonuç metrikleri arasında, ilgili problem ve veri seti düşünüldüğünde, çok büyük bir fark gözlemlenmemiştir.

\begin{figure}[h!]
	\centering
		\includegraphics[width=0.5 \textwidth]{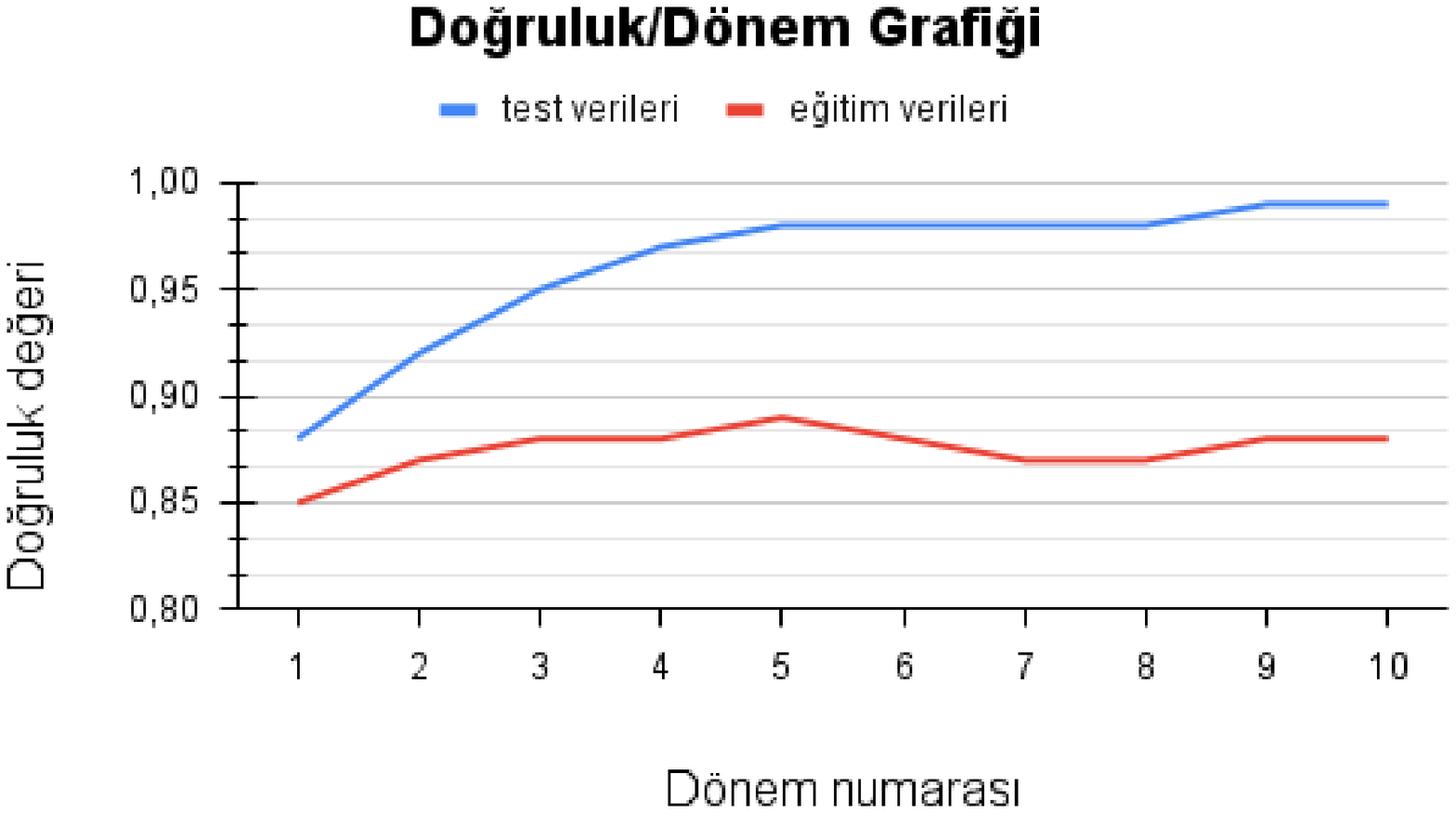}
	    \caption{loodos/bert-base-turkish-cased\cite{loodosbertbaseturkishcased} Modeli 10 Dönem Eğitim Sonuçları}
	\label{img:loodos}
\end{figure}

\begin{figure*}
	\centering
		\includegraphics[width=0.75\textwidth]{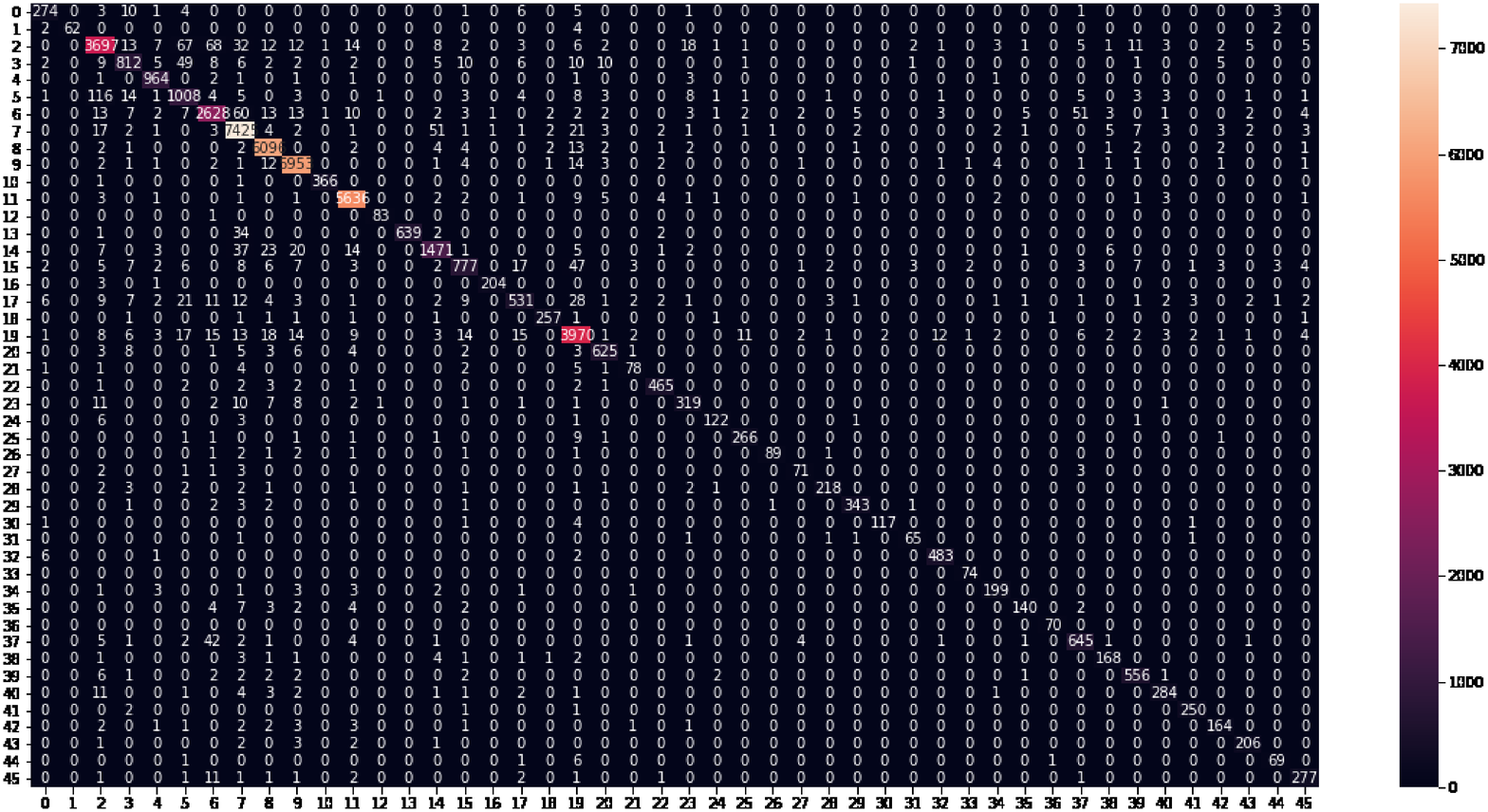}
	\caption{loodos/bert-base-turkish-cased\cite{loodosbertbaseturkishcased} Modeli 3 Dönem Eğitim Sonuçları}
	\label{img:loodos-epoch3matrix}
\end{figure*}

Başarım oranları incelendiğinde en başarılı modelin \%89 doğruluk değeri ve \%82 F1 skoruyla loodos/bert-base-turkish-cased\cite{loodosbertbaseturkishcased} olduğu gözlemlenmiştir. Bu eğitimin çoklu sınıf bazlı başarısı hata dizeyi üzerinden Şekil \ref{img:loodos-epoch3matrix}'de gösterilmiştir. Aynı modelin eğitimi dönem sayısının etkisinin gözlemlenmesi amacıyla 10 dönem ile tekrarlanmıştır. Bulgular sonucunda 3 dönemden sonra  modelin aşırı öğrenmeye gittiği gözlemlenmiştir. Modelin dönemlere göre başarısı doğruluk/dönem grafiğine göre Şekil \ref{img:loodos}'den incelenebilir.

%loodos/bert-base-turkish-cased Modeli 10 Dönem Eğitim Sonuçları grafik

%BERT ve GPT-2 Modelleri Eğitim Sonuçları Grafikleri

%loodos/bert-base-turkish-cased Modeli 3 Dönem Eğitim Sonuçları

\section{Sonuçlar}
Bu çalışmada, önceden eğitilmiş BERT modellerinin sınıflandırma problemi üzerindeki performansları karşılaştırılarak en yüksek başarı seviyesine sahip model belirlenmeye çalışılmış ve bu modeller aynı zamanda kendi eğitttiğimiz GPT-2 modelinin aynı problem özelindeki başarısı ile karşılaştırılmıştır. GPT-2 ile  BERT modelinde elde edilen başarı seviyelerine ulaşılamamıştır. Gerçekleştirilen eğitimler doğrultusunda BERT modelleri arasından Loodos ekibine ait online bloglardan, e-kitaplardan, gazetelerden, Twitter’dan, makalelerden ve Wikipedia sitesi üzerinden toplanan 200GB boyutundaki külliyat ile önceden eğitilmiş, 32000 adet kelime bulunan sözlüğü ve 110 milyon parametreye sahip BERT modelinin en iyi başarım sonuçlarını verdiği gözlemlenmiştir. İlerleyen çalışmalarda, doğruluk değeri \%90'ın altında kalan sınıflar için GPT-2 gibi üretken bir özelliğe sahip model yardımıyla veri artırma (data augmentation) yoluna gidilerek sonuçların iyileştirilebileceği öngörülmektedir.

\section*{B{\footnotesize İ}lg{\footnotesize İ}lend{\footnotesize İ}rme}

Bu çalışma TÜBİTAK TEYDEB 1501 programı kapsamında desteklenmiş olan 3190585 numaralı "Makine Öğrenmesi ile Anlamlı Diyalog Üretebilecek Genel Amaçlı Chatbot Uygulaması" isimli proje kapsamında gerçekleştirilmiştir.

\bibliographystyle{IEEEbib}
\bibliography{referanslar}

\end{document}